\documentclass[nohyperref]{article}

\usepackage{microtype}
\usepackage{graphicx}
\usepackage{booktabs} 

\usepackage{hyperref}

\usepackage{arxiv} 
\usepackage{natbib}

\usepackage{amsmath}
\usepackage{amssymb}
\usepackage{mathtools}
\usepackage{amsthm}

\usepackage{mathtools}
\usepackage{pifont}
\usepackage{amsfonts}
\usepackage{makecell}
\usepackage{placeins}

\usepackage{subcaption}
\usepackage{multirow}

\usepackage[capitalize,noabbrev]{cleveref}

\theoremstyle{plain}

\theoremstyle{definition}

\theoremstyle{remark}

\usepackage[textsize=tiny]{todonotes}

\date{}
\title{Uncertainty in Contrastive Learning: On the Predictability of Downstream Performance}

\author{
  Shervin Ardeshir\\
  Netflix\\
  \texttt{shervina@netflix.com} \\
   \And
  Navid Azizan\\
  Massachusetts Institute of Technology\\
  \texttt{azizan@mit.edu} \\
}

\begin{document}
\maketitle

\begin{abstract}
The superior performance of some of today's state-of-the-art deep learning models is to some extent owed to extensive (self-)supervised \emph{contrastive} pretraining on large-scale datasets. In contrastive learning, the network is presented with pairs of positive (similar) and negative (dissimilar) datapoints and is trained to find an embedding vector for each datapoint, i.e., a representation, which can be further fine-tuned for various downstream tasks. In order to safely deploy these models in critical decision-making systems, it is crucial to equip them with a measure of their \emph{uncertainty} or \emph{reliability}. However, due to the pairwise nature of training a contrastive model, and the lack of absolute labels on the output (an abstract embedding vector), adapting conventional uncertainty estimation techniques to such models is non-trivial. 
In this work, we study whether the uncertainty of such a representation can be quantified for a single datapoint in a meaningful way. In other words, we explore if the downstream performance on a given datapoint is predictable, directly from its pre-trained embedding.
We show that this goal can be achieved by directly estimating the distribution of the training data in the embedding space and accounting for the local consistency of the representations. Our experiments show that this notion of uncertainty for an embedding vector often strongly correlates with its downstream accuracy.
\end{abstract}

\section{Introduction}
Uncertainty estimation is an imperative for safe deployment of deep neural networks in critical decision-making systems. While deep learning approaches are capable of finding useful representations that have demonstrably enabled breakthroughs in a wide variety of tasks, one cannot wishfully assume that their predictions will always be accurate when queried on various inputs. There have been many examples of these systems making wrong predictions, which in some cases have led to fatal accidents \citep{NTSB,varshney2017safety} and unacceptable errors \citep{guynn2015google}. Many such failures may be prevented if the system could supplement its predictions with a level of uncertainty or confidence in those predictions \citep{dietterich2017steps}, which is crucial for building societal trust in such systems. Besides safety, uncertainty estimation is also needed as a part of certain learning algorithms \citep{hullermeier2021aleatoric}, e.g., for uncertainty reduction in active learning \citep{aggarwal2014active}.

To that end, a large body of work on uncertainty estimation for deep neural networks has emerged over the past few years (see, e.g., \citep{blundell2015weight,gal2016dropout,lakshminarayanan2017simple,loquercio2020general,sharma2021sketching, osband2021epistemic}). While there are many sources of uncertainty, in Bayesian modeling, they are often categorized into two types: \emph{aleatoric} and \emph{epistemic}\footnote{Aleatoric uncertainty relates to chance (Latin: \emph{alea} $\leftrightarrow$ dice) and epistemic uncertainty relates to knowledge (ancient Greek: \emph{episteme} $\leftrightarrow$ knowledge) \cite{osband2021epistemic}} \citep{KIUREGHIAN2009105}. Distinguishing between these two types of uncertainties in deep learning has been recently advocated for in the literature \citep{kendall2017uncertainties}. In particular, aleatoric uncertainty refers to the noise inherent in the observations, while epistemic uncertainty captures the uncertainty in the model. Aleatoric uncertainty cannot be reduced even if more data is collected (e.g., sensor noise), while epistemic uncertainty, which accounts for the model's ignorance, can, in principle, be reduced with more data \citep{hullermeier2021aleatoric,kendall2017uncertainties}.

Contrastive learning is a powerful approach to representation learning, which is responsible for the success of many state-of-the-art deep neural networks \citep{simclr,supcon,wu2018unsupervised,henaff2020data,oord2018representation,tian2020contrastive,hjelm2018learning,he2020momentum}. During a typical training regime of a contrastive model, pairs of datapoints are provided as positives or negatives; the contrastive objective then aims to find a data representation in which the positive pairs ``attract,'' (i.e., fall close to each other with an appropriate notion of distance) and the negative pairs ``repulse'' each other in the embedding space. This approach tends to yield rich and robust representations of the data, which may be further used or fine-tuned for downstream tasks. Depending on the availability of labeled data, the contrastive pre-training phase may be performed in a supervised \citep{supcon} or self-supervised \citep{simclr} fashion.

Despite the recent developments in uncertainty quantification, the majority of the literature has been focused on supervised settings, in which a single input is mapped to an absolute target value. Such approaches are not readily applicable to contrastive models, as the model's prediction on a single datapoint is an abstract embedding vector. Nonetheless, such approaches are applicable for measuring the (un)certainty of the model on a \textit{pair} of datapoints, treating contrastive models as binary classifiers. To that end, a Bayesian metric learning framework was proposed by \cite{wang2017robust} to construct a robust estimation of the distance, given a pair of datapoints. Measuring the uncertainty of a metric learning model, given a pair of datapoints, was further explored by \cite{qian2018large}. A relatively close work to ours is that of \cite{oh2018modeling}, in which the aleatoric embedding uncertainty is evaluated in terms of instance retrieval, which is the objective the model is directly trained for. Unlike these works, we aim to measure the uncertainty/reliability of the embedding for \textit{a single datapoint}, in such a way that is predictive of its \emph{downstream performance}. The closest works to ours are recent works of \cite{zhang2021temperature, wu2020simple}, both of which tackle the problem of uncertainty estimation in the case of a contrastive objective. The key difference between our work and theirs is two folds: First, unlike \cite{zhang2021temperature, wu2020simple}, in our setup, a pre-trained model is given to us as a black box, for which the uncertainty of a datapoint should be estimated. In the aforementioned work, the model is being pre-trained using a modified objective function, to incorporate the notion of uncertainty. Our work, on the contrary, is a post-processing step on a pre-trained black-box contrastive model. Second, in our work, the notion of certainty for a datapoint is defined as its downstream performance, as opposed to the work of \cite{zhang2021temperature, wu2020simple}, in which a small variance of the embedding is indicative of certainty.

\begin{figure}
    \centering
    \includegraphics[width=0.7\linewidth]{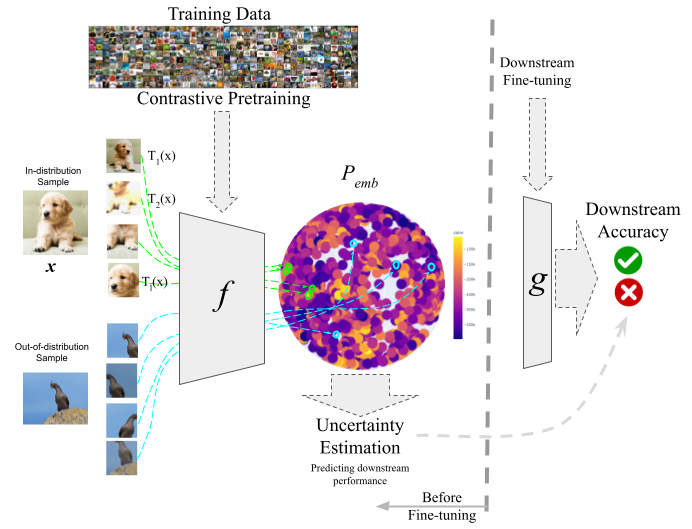}
    \caption{We are given a contrastive model $f$, which is pre-trained in a supervised or self-supervised manner on a training dataset. Given a test image $x$, we measure the reliability of its resulting embedding $f(x)$ using notions such as embedding-variance (given different augmentations), and the distribution of the training data embeddings, to quantify this reliability. We show this reliability is not only capable of detecting out of distribution samples (epistemic uncertainty), but also predictive of the performance on a datapoint in a downstream task. }
    \label{fig:teaser}
\end{figure}

In this paper, we explore the possibility of capturing the reliability of an embedding resulting from a (black box) pre-trained contrastive model, in terms of predicting downstream performance.
Figure \ref{fig:teaser} shows an overview of our setup. A back-box model $f$ is pre-trained with a contrastive objective, on a training dataset, resulting in an embedding vector representing each datapoint. The goal is to study whether there are any notions of uncertainty for an embedding vector which is indicative of its reliability, \textit{i.e.} how it would later perform downstream. Given the non-triviality of predicting downstream performance solely from a pre-trained embedding vector, we explore the possibility of such prediction using a few intuitive measures. To that end, given an input and a pre-trained model, we measure the reliability of the resulting embedding in three aspects: (1) \textit{How certain the model is about the location of an embedding vector}. This is computed by introducing variations to the input datapoint and measuring variations in its embedding vector. (2) \textit{How familiar the model is with that area of the embedding space}. In other words, has the model seen training examples with similar embeddings. This notion is computed by directly estimating the distribution of the embedding vectors of the training data. (3) \textit{How well does the model perform in that region of the embedding space}. This is measured by calculating the local retrieval performance of the model. We study whether these intuitive notions meaningfully correlate with the downstream performance on a given input.

\section{Framework and Proposed Method}
In this section, we describe our framework and how our various notions of uncertainty/reliability are constructed. Let us consider a model $f:\mathbb{R}^n\to\mathcal{S}^{m-1}$, which maps an $n$-dimensional input datapoint (e.g., an image) $x$ to the $\ell_2$-normalized $m$-dimensional feature vector $f(x)$ (on the unit hypersphere). Given the model $f$, we aim to measure the reliability of the embedding vector $f(x)$ for any given input $x$. As discussed earlier, we do so based on quantifying the uncertainty in the location of the point in the embedding space as well as the consistency of the model's prediction in that region. 

\subsection{Per-Sample Feature Variation: $\delta$}
\label{sec:framework_feat_var}

The first notion of uncertainty that we define aims to capture \textit{how certain the model is about the location of an embedding vector}. Given a set of variations/transformations\footnote{These are often referred to as data ``augmentation'' techniques because they are used for augmenting the dataset. Geometric transformations, flipping, color modification, cropping, rotation, noise injection and random erasing are among the common ones \citep{shorten2019survey}.} $\{ T_1, T_2, \dots, T_l\}$ used in the training of the contrastive model, we measure the variation across $\{ z_1, z_2, \dots, z_l\}$, where $z_i=f(T_i(x))$ is the embedding vector corresponding to the $i$-th transformation of the input $x$. More specifically, we define $\delta({x})$ as the sum of the variances for different dimensions, i.e., the trace of the sample covariance matrix for the observation vectors $\{ z_1, z_2, \dots, z_l\}$. Our experiments show that this simple quantity often meaningfully predicts the reliability of the embedding, as measured by the performance in a downstream task. An important characteristic of this metric is that it {\textit{does not require access to the training data}} and would work on any black-box model. Note that the underlying assumption here is that the downstream task is invariant to the pre-training data transformations (augmentations).

\subsection{Embedding Distribution Estimation}
\label{sec:framework_distribution_estimation}
This notion is based on estimating the distribution of the embedding, which we refer to as $p_\mathrm{emb}$. This probability distribution captures the two key features of {\textit{density}} and {\textit{consistency}} over the unit hypersphere embedding space.

\begin{table}
\centering
\begin{tabular}{c||c|c|c|c|}
  & Model &  \makecell{Training\\Inputs} & \makecell{Training\\Labels} & \makecell{Downstream\\Classifier}\\ [0.5ex] 
 \hline\hline
 Per-Sample Feature Variation: $\delta$  (\ref{sec:framework_feat_var}) & \ding{51} & $\times$ & $\times$ & $\times$ \\
 \hline
 Embedding Density: $p_\mathrm{emb}$ (\ref{sec:framework_desnity}) & \ding{51} & \ding{51} & $\times$ & $\times$ \\
 \hline
 Embedding Consistency: $p_\text{emb-ens}$ (\ref{sec:framework_consistency}) & \ding{51} & \ding{51} & $\times$ & $\times$ \\
 \hline
 Ensembled Embedding Density: $p^{k,\tau}_\mathrm{emb}$ (\ref{sec:framework_ensemble}) & \ding{51} & \ding{51} & \ding{51} & $\times$ \\
 \hline
 Ensembled Embedding Consistency: $p^{k,\tau}_\text{emb-ens}$ (\ref{sec:framework_ensemble}) & \ding{51} & \ding{51} & \ding{51} & $\times$ \\
 \hline\hline
 Entropy (\ref{sec:baselines}) & \ding{51} & \ding{51} & \ding{51} & \ding{51} \\
 \hline
 Max Score (\ref{sec:baselines}) & \ding{51} & \ding{51} & \ding{51} & \ding{51} \\
 \hline
\end{tabular}
\caption{Our different uncertainty measures make different assumptions on access to the model and the training data. A \ding{51} indicates requiring access. Note that entropy and max score require access to the downstream classifier and are thus not applicable in our setting.}
\label{tab:requirements}
\end{table}

\subsubsection{Density: $p_\mathrm{emb}$}
\label{sec:framework_desnity}
The density of the embedding space at a point $z$ would intuitively capture {\textit{how much data has the model observed}} around $z$ during training, which is the transformation of the training data distribution under $f$. To estimate this distribution, we fit a Gaussian\footnote{To be precise, one has to use a Fisher–Bingham (or Kent) distribution \citep{jupp2009directional,kent1982fisher} over the $(m-1)$-sphere, which is the analogue of a Gaussian on hypersphere.} mixture model (GMM) to the $m$-dimensional embeddings of the training data. Computing this density function requires access to the pre-trained model and an \emph{unsupervised} training dataset, i.e., only the input datapoints and not the labels.      

\subsubsection{Consistency: $p^{k,\tau}_\mathrm{emb}$}
\label{sec:framework_consistency}
The consistency of the model at $z$ measures whether the training datapoints mapped closest to $z$ have consistent labels. This notion would capture \textit{how accurate the model is} at $z$, based on the fact that a more accurate contrastive model should have a more pure local correspondence. Note that unlike the density-only distribution mentioned above, estimating this distribution requires access to both training data and training labels (correspondences). For each training datapoint, we calculate the fraction of its $k$ nearest neighbors ($k$-NN) in the embedding space whose class labels are consistent with that datapoint. We then filter out the datapoints based on their $k$-NN accuracy with a threshold $\tau$, and fit a Gaussian mixture model to the datapoints whose $k$-NN consistency is above the threshold $\tau$. We denote this distribution by $p^{k,\tau}_\mathrm{emb}$. This notion would require access to the model and a \emph{supervised} training dataset, and is thus only applicable to the supervised contrastive learning setup \citep{supcon}. 

It is worth noting that setting the threshold $\tau$ to zero yields $p^{k,0}_\mathrm{emb}(\cdot)=p_\mathrm{emb}(\cdot)$, which would solely capture the density of each datapoint in the training data. 

Intuitively, the two notions defined above could lead to the following scenarios:
\begin{itemize}
\item {High $p_\mathrm{emb}(z)$ and high $p_\mathrm{emb}^{k,\tau}(z)$:} The model has seen many consistent examples like $z$ (low uncertainty).
\item {High $p_\mathrm{emb}(z)$ and low $p^{k,\tau}_\mathrm{emb}(z)$:} The model has seen samples similar to $z$ during training, but has not been consistent for them. This could be due to the fact that these are hard examples, thus implying low epistemic uncertainty but high aleatoric uncertainty.
\item {Low $p_\mathrm{emb}(z)$:} The model has not seen samples similar to $z$. This implies the sample is likely out-of-distribution with respect to the training set, and thus has a high epistemic uncertainty.
\end{itemize}

\subsection{Per-Sample Feature Variance + Embedding Distribution: $p^{k,\tau}_\text{emb-ens}$}
\label{sec:framework_ensemble}
One could also combine the two notions of per-sample variance and embedding distribution, which has the interpretation of a stochastic embedding \citep{wang2020understanding}. More specifically, we have an ensemble of probabilities through the $l$ transformations $\{ T_1, T_2, ..., T_l \}$, and using the law of total probability we have
$$p^{k,\tau}_\text{emb-ens}=\sum_{i=1}^l p^{k,\tau}_\mathrm{emb}(f(T_i(x)))p(T_i) = \frac1l \sum_{i=1}^l p^{k,\tau}_\mathrm{emb}(f(T_i(x))).$$

The measures mentioned above have different requirements, ranging from access to the black-box model only (feature-variation measure), to requiring access to a fully supervised training dataset (consistency measure). Table~\ref{tab:requirements} summarizes the requirements for each measure. Note that the last two measures (entropy and max score), which are explained in Section~\ref{sec:experiments}, require the full observation of the downstream task and are solely defined as a baseline.

\section{Experimental Results}
\label{sec:experiments}
We pre-train self-supervised (SimCLR) \citep{simclr} and supervised (SupCon) \citep{supcon} contrastive models with ResNet18 \citep{resnet} backbones, and on the training set of CIFAR10 or CIFAR100 \citep{cifar} datasets. We then perform inference on their test sets, alongside test sets of CUBS2011 \citep{cubs2011} and SVHN \citep{svhn} as other out-of-distribution datasets. We follow the pre-training and linear fine-tuning protocols in accordance to \cite{supcon}. 

\begin{table}
\centering
\begin{tabular}{cc||c|c|c|c|}
  && Aleatoric &  Epistemic & Overall\\ [0.5ex] 
 \hline\hline
  \multirow{2}{*}{In-distribution} &  Correct & 0 & 0 & 0 \\
 \cline{2-5}
  & Incorrect & 1 & 0 & 1 \\
 \hline
 Out-of-distribution & & - & 1 & 1 \\
 \hline
\end{tabular}
\caption{This table shows how the ground-truth labels for evaluating each of our notions of uncertainty are defined based on in- vs. out-of-distribution and the correctness of the downstream classifier. Aleatoric uncertainty evaluates whether the measure correlates with datapoint difficulty (downstream failure) on the in-distribution samples. Epistemic uncertainty measures whether unfamiliar (out-of-distribution) samples are distinguishable from familiar (in-distribution) datapoints. Overall uncertainty measures the retrieval of correctly classified in-distribution datapoints.}
\label{tab:gt_labels}
\end{table}

\subsection{Uncertainty Measures}
\label{sec:baselines}
As discussed earlier, our different uncertainty measures make different assumptions about access to the data and models, according to which we categorize them into several groups.

\textbf{Pre-trained model only.} The feature-variation measure ($\delta$), described in Section~\ref{sec:framework_feat_var}, only requires access to the pre-trained model. Our quantitative results indicate that this very simple approach (directly applicable at inference time) already allows for measuring uncertainty in many scenarios.

\textbf{Pre-trained model + unsupervised training data.} $p_\mathrm{emb}$ and $p_\text{emb-ens}$ (described in Section~\ref{sec:framework_distribution_estimation})
would require access to the model and the training dataset without supervision (i.e., training images only), making them applicable to both self-supervised and supervised setups. The training data is only used for a single forward pass.

\textbf{Pre-trained model + supervised training data.} $p^{k,\tau}_\mathrm{emb}$ and $p^{k,\tau}_\text{emb-ens}$ require access to the labeled training dataset, as they incorporate the consistency notion mentioned in Section~\ref{sec:framework_consistency}. Thus, these measures are only applicable to the supervised contrastive \citep{supcon} setup. In our experiments, we define consistency as having top 1\% $k$-NN accuracy threshold of 50\%. We also study the impact of $k$ (the number of neighbors) and $\tau$ (the threshold) on different types of uncertainty estimation metrics in Section~\ref{sec:ablation}.

\textbf{Fine-tuned model.}
The following two measures are not computable in our scenario, as they are only measurable \emph{after} a downstream classifier is fine-tuned on the pre-trained features. Thus, our approach would not be comparable to these measures. Regardless, we report the measurements to put our quantitative measurements in context. Our experiments indicate that in some scenarios, our measures achieve competitive, or sometimes even slightly better, performance compared with these measures.\\  
\textit{Entropy}: Entropy of a classifier is often used as a measure of uncertainty. We measure the entropy of the downstream fine-tuned classifier on each sample and use that as a measure of certainty.  \\
\textit{Max score}: The maximum score (confidence) of the downstream classifier is used as a measure of certainty.

A summary of the requirements for each of the measures is provided in Table~\ref{tab:requirements}.

\begin{table*}[t]
\resizebox{\textwidth}{!}{%
\begin{tabular}{|c|c||c||c|c||c|c||c|c|c|} 
 \hline
 Setup & Dataset &  $\delta$ & $p_\mathrm{emb}$ & $p_\text{emb-ens}$ & $p^{k,\tau}_\mathrm{emb}$ & $p^{k,\tau}_\text{emb-ens}$ & Entropy & Max score\\ [0.5ex] 
 \hline\hline
SimCLR & CIFAR10 & 0.652 & 0.702 $\pm$ 0.023 & \textbf{0.719 $\pm$ 0.004} & - & - & 0.883 & 0.836\\
SimCLR & CIFAR100 & \textbf{0.647} & 0.559 $\pm$ 0.009 & 0.564 $\pm$ 0.017 & - & - & 0.816 & 0.761\\
SupCon & CIFAR10 & 0.830 & 0.805 $\pm$ 0.025 & 0.858 $\pm$ 0.004 & 0.808 $\pm$ 0.026 & \textbf{0.862 $\pm$ 0.002} & 0.916 & 0.892\\
SupCon & CIFAR100 & \textbf{0.766} & 0.735 $\pm$ 0.017 & 0.764 $\pm$ 0.004 & 0.720 $\pm$ 0.017 & 0.743 $\pm$ 0.002 & 0.879 & 0.852\\
 \hline
\end{tabular}
}
\caption{Aleatoric uncertainty (quantified based on AUROC as described in \ref{sec:experiments_aleatoric}). Predicting performance on each datapoint in the downstream task of image classification. As it can be observed, all of our notions meaningfully capture sample difficulty as measured by correctness on in-distribution datapoints.}
\label{tab:aleatoric}
\end{table*}

\begin{table*}[t]
\resizebox{\textwidth}{!}{%
\begin{tabular}{|c|c|c||c||c|c||c|c||c|c|} 
 \hline
 {Setup} & {In-dist} & {Out-of-dist} & {$\delta$} & {$p_\mathrm{emb}$} & {$p_\text{emb-ens}$} & {$p^{k,\tau}_\mathrm{emb}$} & {$p^{k,\tau}_\text{emb-ens}$} & {Entropy} & {Max score}\\ [0.5ex] 
 \hline\hline
SimCLR & CIFAR10 & CUBS2011 & \textbf{0.766} & 0.59 $\pm$ 0.004 & 0.602 $\pm$ 0.074 & - & - & 0.689 & 0.745\\
SimCLR & CIFAR10 & SVHN & 0.393 & 0.960 $\pm$ 0.000 & \textbf{0.975 $\pm$ 0.018} & - & - & 0.890 & 0.918\\
SimCLR & CIFAR10 & CIFAR100 & 0.645 & 0.773 $\pm$ 0.003 & \textbf{0.793 $\pm$ 0.035} & - & - & 0.851 & 0.858\\
SimCLR & CIFAR100 & CUBS2011 & \textbf{0.783} & 0.598 $\pm$ 0.0032 & 0.608 $\pm$ 0.020 & - & - & 0.775 & 0.783\\
SimCLR & CIFAR100 & SVHN & 0.365 & 0.810 $\pm$ 0.002 & \textbf{0.846 $\pm$ 0.022} & - & - & 0.761 & 0.789\\
SimCLR & CIFAR100 & CIFAR10 & \textbf{0.610} & 0.515 $\pm$ 0.0023 & 0.516 $\pm$ 0.010 & - & - & 0.692 & 0.670\\
SupCon & CIFAR10 & CUBS2011 & 0.580 & 0.644 $\pm$ 0.005 & \textbf{0.660 $\pm$ 0.031} & 0.640 $\pm$ 0.006 & 0.655 $\pm$ 0.022 & 0.671 & 0.690\\
SupCon & CIFAR10 & SVHN & 0.548 & 0.977 $\pm$ 0.003 & \textbf{0.995 $\pm$ 0.000} & 0.976 $\pm$ 0.003 & 0.995 $\pm$ 0.001 & 0.962 & 0.964\\
SupCon & CIFAR10 & CIFAR100 & 0.765 & 0.878 $\pm$ 0.003 & \textbf{0.918 $\pm$ 0.002} & 0.877 $\pm$ 0.003 & 0.916 $\pm$ 0.002 & 0.903 & 0.900\\
SupCon & CIFAR100 & CUBS2011 & 0.853 & 0.727 $\pm$ 0.006 & \textbf{0.762 $\pm$ 0.084} & 0.718 $\pm$ 0.004 & 0.747 $\pm$ 0.026 & 0.877 & 0.884\\
SupCon & CIFAR100 & SVHN & 0.546 & 0.904 $\pm$ 0.003 & \textbf{0.940 $\pm$ 0.015} & 0.867 $\pm$ 0.002 & 0.903 $\pm$ 0.017 & 0.845 & 0.852\\
SupCon & CIFAR100 & CIFAR10 & \textbf{0.720} & 0.667 $\pm$ 0.005 & 0.689 $\pm$ 0.016 & 0.644 $\pm$ 0.002 & 0.661 $\pm$ 0.028 & 0.739 & 0.723\\
\hline
\end{tabular}
}
\caption{Epistemic uncertainty (quantified based on AUROC as described in \ref{sec:experiments_epistemic}) is evaluated by measuring performance in terms of out-of-distribution detection. As expected, $p_\mathrm{emb}$, which aims to directly estimate the embedding distribution, outperforms other measures in most instances.}
\label{tab:epistemic}
\end{table*}

\begin{table*}[t]
\resizebox{\textwidth}{!}{%
\begin{tabular}{|c|c|c||c||c|c||c|c||c|c|} 
 \hline
 {Setup} & {In-dist} & {Out-of-dist} & {$\delta$} & {$p_\mathrm{emb}$} & {$p_\text{emb-ens}$} & {$p^{k,\tau}_\mathrm{emb}$} & {$p^{k,\tau}_\text{emb-ens}$} & {Entropy} & {Max score}\\ [0.5ex] 
 \hline\hline
SimCLR & CIFAR10 & CUBS2011 & \textbf{0.768} & 0.627 $\pm$ 0.006 & 0.640 $\pm$ 0.071 & - & - & 0.765 & 0.802\\
SimCLR & CIFAR10 & SVHN & 0.424 & 0.956 $\pm$ 0.001 & \textbf{0.970 $\pm$ 0.017} & - & - & 0.926 & 0.945\\
SimCLR & CIFAR10 & CIFAR100 & 0.666 & 0.785 $\pm$ 0.003 & \textbf{0.805 $\pm$ 0.036} & - & - & 0.896 & 0.894\\
SimCLR & CIFAR100 & CUBS2011 & \textbf{0.781} & 0.600 $\pm$ 0.006 & 0.609 $\pm$ 0.022 & - & - & 0.862 & 0.848\\
SimCLR & CIFAR100 & SVHN & 0.448 & 0.798 $\pm$ 0.005 & \textbf{0.829 $\pm$ 0.019} & - & - & 0.865 & 0.875\\
SimCLR & CIFAR100 & CIFAR10 & \textbf{0.668} & 0.539 $\pm$ 0.003 & 0.541 $\pm$ 0.010 & - & - & 0.810 & 0.768\\
SupCon & CIFAR10 & CUBS2011 & 0.616 & 0.675 $\pm$ 0.004 & \textbf{0.697 $\pm$ 0.028} & 0.671 $\pm$ 0.004 & 0.692 $\pm$ 0.021 & 0.713 & 0.729\\
SupCon & CIFAR10 & SVHN & 0.572 & 0.979 $\pm$ 0.003 & \textbf{0.995 $\pm$ 0.000} & 0.978 $\pm$ 0.003 & 0.995 $\pm$ 0.001 & 0.975 & 0.976\\
SupCon & CIFAR10 & CIFAR100 & 0.788 & 0.891 $\pm$ 0.003 & \textbf{0.931 $\pm$ 0.002} & 0.890 $\pm$ 0.003 & 0.930 $\pm$ 0.002 & 0.927 & 0.922\\
SupCon & CIFAR100 & CUBS2011 & 0.877 & 0.773 $\pm$ 0.012 & \textbf{0.811 $\pm$ 0.061} & 0.760 $\pm$ 0.009 & 0.790 $\pm$ 0.017 & 0.935 & 0.933\\
SupCon & CIFAR100 & SVHN & 0.645 & 0.916 $\pm$ 0.005 & \textbf{0.947 $\pm$ 0.009} & 0.8820 $\pm$ 0.005 & 0.912 $\pm$ 0.011 & 0.926 & 0.926\\
SupCon & CIFAR100 & CIFAR10 & \textbf{0.794} & 0.727 $\pm$ 0.010 & 0.756 $\pm$ 0.014 & 0.703 $\pm$ 0.007 & 0.725 $\pm$ 0.025 & 0.854 & 0.828\\
\hline
\end{tabular}
}
\caption{Overall uncertainty (quantified based on AUROC as described in \ref{sec:overall_unc}). Similar to epistemic uncertainty, $p_\mathrm{emb}$ seem to often yield better quantitative metrics, especially in scenarios in which the feature variation measure seems to fail, which shows the two notions are complementary.}
\label{tab:overall}
\end{table*}

\subsection{Evaluation}
\label{sec:evaluation_metrics}
We evaluate different notions of uncertainty to cover different aspects of downstream predictability. To evaluate the different metrics, we treat them as retrieval instances and compute their AUROC (Area Under the Receiver Operating Characteristic curve). The ground-truth label of the retrieval instance could be derived as a function of the downstream accuracy of a datapoint and whether the model has been exposed to the datapoint's semantic class during pre-training. In what follows, we discuss the details of this evaluation for each uncertainty notion. Table~\ref{tab:gt_labels} summarizes what each uncertainty notion is capturing.

\begin{figure}[b]
     \centering
     \begin{subfigure}{.48\linewidth}
         \includegraphics[width=\linewidth]{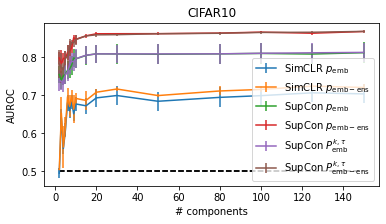}
     \end{subfigure}
     \begin{subfigure}{.48\linewidth}
         \includegraphics[width=\linewidth]{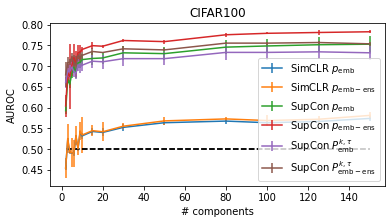}
     \end{subfigure}
     \caption{The effect of number of GMM components on our aleatoric uncertainty measures. We generally observe improvements with larger $n_\text{comp}$. However, for $n_\text{comp}>20$, the improvements seem rather marginal and the results seem stable across all models and datasets.}
     \label{fig:aleatoric_ncomp}
\end{figure}

\subsubsection{Aleatoric Uncertainty} 
\label{sec:experiments_aleatoric}
Aleatoric uncertainty is often defined as the ``noise inherent in the data,'' which leads to difficulty of understanding a sample datapoint. We use downstream performance of a datapoint as a proxy for measuring its difficulty. 
To quantify such notion, we evaluate our proposed uncertainty measures on in-distribution test-set datapoints, and in terms of their capability in retrieving samples which are correctly classified in a downstream linear classifier. Table~\ref{tab:aleatoric} shows this metric for our different uncertainty measures. 

\subsubsection{Epistemic Uncertainty}
\label{sec:experiments_epistemic}
We evaluate our uncertainty estimation measures on images from the in-distribution (pre-training dataset) and an out-of-distribution dataset and quantify their performance in terms of retrieving the in-distribution embeddings. In other words, a model pre-trained (supervised or self-supervised) on the training set of dataset A is fed test datapoints from datasets A and B. Then, the effectiveness of the uncertainty measures are evaluated in terms of distinguishing datapoints of dataset A from those of B. Table~\ref{tab:epistemic} contains the performance of our different measures on this task. It can be observed that in most cases, $p_\mathrm{emb}$ has the best performance, whose definition is also more consistent with out-of-distribution detection tasks, as it directly estimates the embedding distribution that comes from the training data. Another observation would be the failure of the feature variation measure in detecting out-of-distribution samples of SVHN in the self-supervised setups. We hypothesize this could be due to the fact that SVHN is a less diverse dataset, which results in its images being mapped close to one another in a CIFAR10 or CIFAR100 pre-trained model. As a result, feature variation would not be a good notion for distinguishing such samples. On the other hand, the probability-based measures result in very high AUROC scores, alluding that these measures capture complementary notions of reliability. Another explanation could be that measuring the effect of data transformation could be interpreted as mainly a notion of \textit{data uncertainty}. This would suggest that this measure is a better fit for aleatoric uncertainty estimation.

\begin{figure}[t]
     \centering
     \begin{subfigure}{.48\linewidth}
         \includegraphics[width=\linewidth]{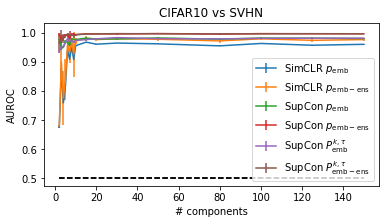}
     \end{subfigure}
     \begin{subfigure}{.48\textwidth}
         \includegraphics[width=\linewidth]{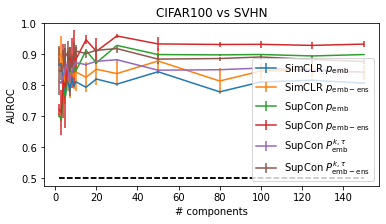}
     \end{subfigure}
     \begin{subfigure}{.48\textwidth}
         \includegraphics[width=\linewidth]{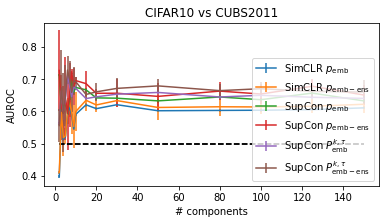}
     \end{subfigure}
     \begin{subfigure}{.48\textwidth}
         \includegraphics[width=\linewidth]{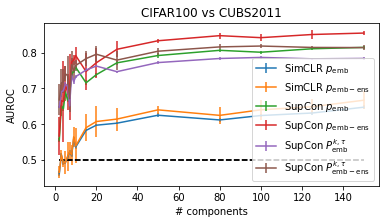}
     \end{subfigure}
     \begin{subfigure}{.48\textwidth}
         \includegraphics[width=\linewidth]{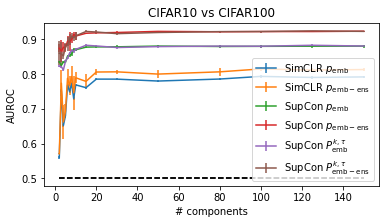}
     \end{subfigure}
        \label{fig:three graphs}
     \begin{subfigure}{.48\textwidth}
         \includegraphics[width=\linewidth]{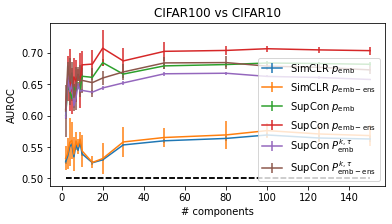}
     \end{subfigure}
        \caption{The effect of number of GMM components on different epistemic uncertainty measures. The titles specify the in-distribution (pre-training) datasets vs. the out-of-distribiution datasets. Similar to the other types of uncertainty, we generally observe improvements with larger $n_\text{comp}$. However the improvements are rather marginal for higher $n_\text{comp}$, and the results seem stable across all models and datasets.}
        \label{fig:epistemic_ncomp}
\end{figure}

On another note, we observe that our probability measures have lower discriminative power distinguishing between CIFAR100 and CIFAR10, as opposed to between CIFAR examples and non-CIFAR examples. This observation is consistent across both datasets, and across both supervised (SupCon) and self-supervised (SimCLR) setups. Also, detecting CIFAR10 samples as out-of-distribution, given a CIFAR100 pre-trained $p_\mathrm{emb}$ model, is noticeably more difficult than distinguishing CIFAR100 samples using a CIFAR10 pre-trained $p_\mathrm{emb}$. 

Here we followed standard practice for evaluating epistemic uncertainty. However, we argue that the assumption of all the datapoints in the test set of dataset A being ``in-distribution'' to the model may not necessarily hold. To address that, we define the following alternative, which captures such nuances.

\subsubsection{Overall Uncertainty}
\label{sec:overall_unc}
Here we introduce a hybrid definition of uncertainty, taking into account both aleatoric and epistemic uncertainties. Given a model trained on dataset A, we evaluate its uncertainty measures on datasets A and B. We then evaluate how well the uncertainty measure retrieves datapoints in A which are \textit{correctly classified by the downstream classifier}. In other words, the model should not be certain about all the datapoints in A, but only the ones that are going to be correctly classified downstream. The quantitative measures using this metric are reported in Table~\ref{tab:overall}. Comparing values in this table with their corresponding values in the epistemic uncertainty evaluation (Table~\ref{tab:epistemic}), we generally observe higher values across all measures. 

\subsection{Ablation Study}
\label{sec:ablation}
In this section, we analyze the effect of different parameters on the performance of our approach.

\begin{figure}[t]
    \centering
    \includegraphics[width=.8\textwidth]{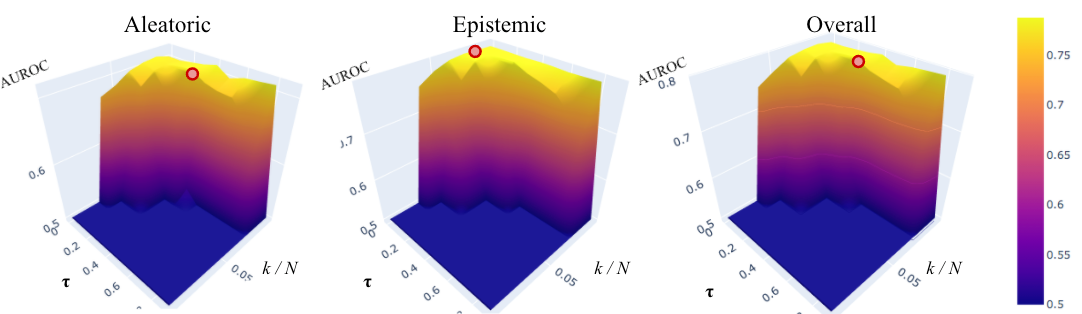}
    \caption{Observing the effect of $k$ (number of neighbors) and $\tau$ (threshold) in the performance of $p_\mathrm{emb}^{k,\tau}$ shows a diminishing return on using  higher accuracy thresholds, as it leads to pruning a large number of datapoints. For epistemic uncertainty (center), we observe the peak of performance to be at $\tau=0$, which corresponds to directly estimating embedding distribution of all of the training set. This is consistent with the fact that epistemic uncertainty is evaluated by detecting out-of-distribution samples. On the other hand, for aleatoric and overall uncertainties, a marginal improvement is achieved at higher thresholds, as the consistency notion does contribute to better downstream performance prediction.}
    \label{fig:knn_and_thr}
\end{figure}

\begin{figure}[t]
    \centering
    \includegraphics[width=.85\textwidth]{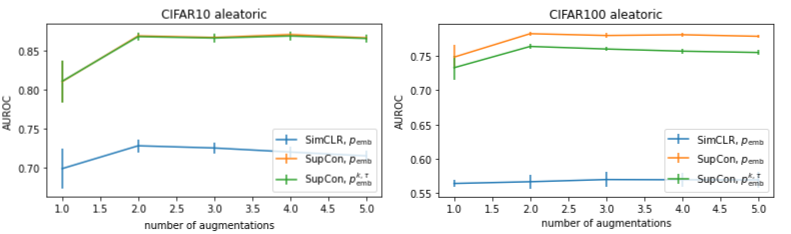}
    \caption{The effect of number of transformations on aleatoric uncertainty estimation. Unlike other variations of uncertainty, we observe no improvement with more than 2 augmentations in most cases.}
    \label{fig:augmentations_aleatoric}
\end{figure}

\textbf{Effect of number of GMM components.}
We evaluate the effect of the number of GMM components ($n_\text{comp}$), by evaluating the metrics, while sweeping $n_\text{comp}$ from 2 to 150 components. Figures~\ref{fig:aleatoric_ncomp} and \ref{fig:epistemic_ncomp} show the aleatoric and epistemic uncertainty measures (y-axis), respectively, using different number of components (x-axis). Interestingly, except for extremely small values ($n_\text{comp}<10$), we observe relatively stable performance across all setups. This observation seems to be consistent across both supervised and self-supervised setups. The overall uncertainty has a very similar trend as the epistemic uncertainty. For the sake of brevity, we relegated the result to the Appendix.


\begin{figure}[t]
    \centering
    \includegraphics[width=.75\textwidth]{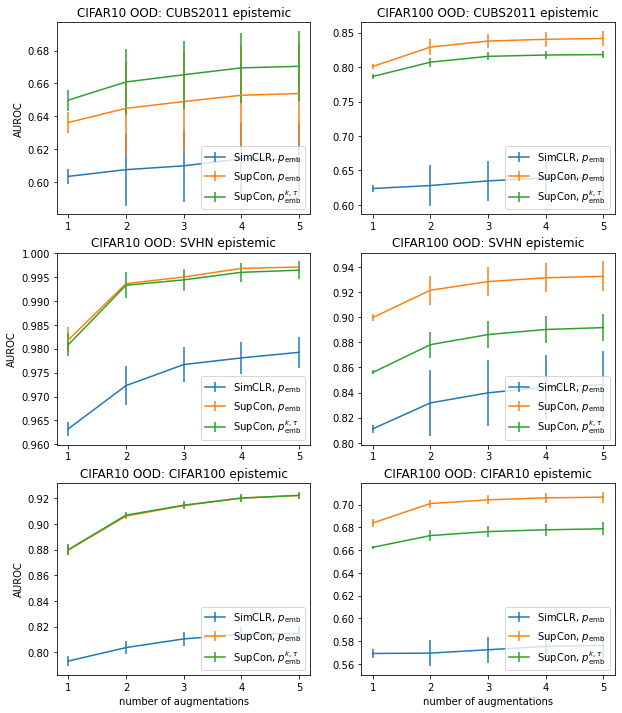}
    \caption{The effect of number of transformations on epistemic uncertainty estimation. As expected, combining likelihoods resulting from more augmentations result in better out-of-distribution detection.}
    \label{fig:augmentations_epistemic}
\end{figure}

\textbf{Effect of threshold and number of nearest neighbors.} Higher thresholds and number of nearest neighbors would result in maintaining highly consistent points, at the cost of losing information (having less remaining datapoints for estimating the GMMs). The diminishing returns of such parameters can be seen in Figure~\ref{fig:knn_and_thr}. It can be observed that peak-performance for epistemic uncertainty (middle), is at threshold of 0 ($p_\mathrm{emb}$), which is consistent with its definition of directly estimating the training data (in-distribution likelihood). On the contrary, consistency does improve the estimation of aleatoric uncertainty, as the peak of the distribution does occur at ($\tau=0.4$, $k/N=0.025$), which means that a consistent datapoint is defined as one whose top 2.5\% of nearest neighbors are more than 40\% consistent with its semantic label. Overall, we observe the effect of consistency to be marginal in our experiments. We hypothesize that such behavior could be due to a highly accurate pre-training, resulting in a high correlation between consistency and density. This correlation could be measured at larger scale and across different setups and datasets (with different accuracies) to validate this hypothesis.

\textbf{Effect of number of transformations.} Figure~\ref{fig:augmentations_aleatoric} shows the effect of number of augmentations on aleatoric uncertainty estimation. It can be observed that after 2 augmentations, the results are relatively stable. On the other hand, for epistemic uncertainty, shown in Figure~\ref{fig:augmentations_epistemic}, more improvement (yet marginal) could be achieved with more augmentations. The overall uncertainty has a very similar behavior to the epistemic uncertainty, shown in Figure~\ref{fig:augmentations_overall}.

\section{Conclusion}
In this paper, we explored the possibility of estimating a reliability/uncertainty measure for the abstract embeddings of contrastive models. We show that our uncertainty measures not only are able to meaningfully detect out-of-distribution samples but also are predictive of performance in downstream tasks.  
We believe that having such notion of reliability/uncertainty can particularly be insightful, e.g., for deciding between different options of pre-trained models, or for deciding on specific sample weighting policies in downstream fine-tuning tasks.

\FloatBarrier
\bibliographystyle{unsrtnat}
\bibliography{references}
\newpage
\onecolumn
\appendix
\section{Appendix}

\textbf{Effect of number of transformations} (for overall uncertainty)

\begin{figure}[h]
    \centering
    \includegraphics[width=.75\textwidth]{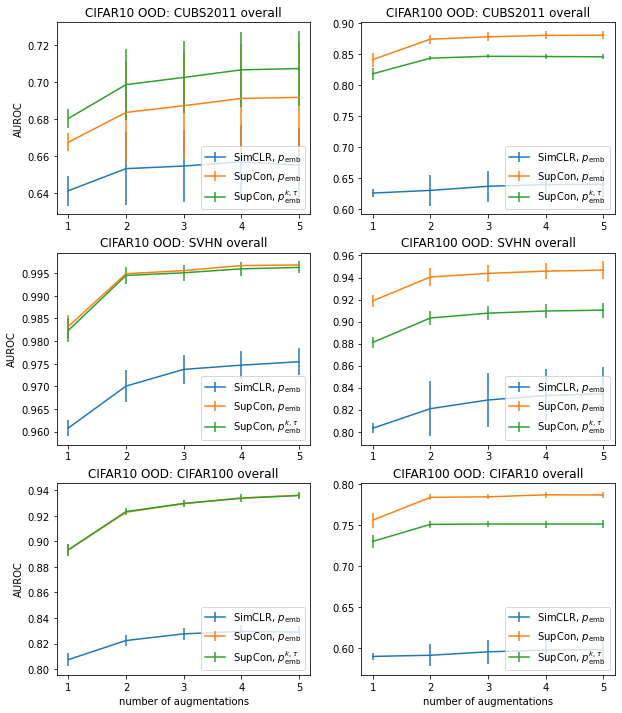}
    \caption{The effect of number of transformations on overall uncertainty. Similar as in the case of epistemic uncertainty, we observe an increase in the performance with more transformations.}
    \label{fig:augmentations_overall}
\end{figure}

As mentioned in Section~\ref{sec:ablation}, we provide the effect of number of data transformations (image augmentations) on the overall uncertainty estimation, shown in Figure~\ref{fig:augmentations_overall}. We observe the trends to be very similar to those of epistemic uncertainty. We observe a monotonically increasing performance as a function of the number of transformations. However, in many scenarios such as SupCon on CIFAR10 vs. SVHN, and CIFAR100 vs. CUBS2011, the amount of improvement becomes marginal beyond 2 augmentations.

\FloatBarrier
\textbf{Effect of number of GMM components} (for overall uncertainty)

\begin{figure}[h]
     \centering
     \begin{subfigure}{.48\textwidth}
         \includegraphics[width=\textwidth]{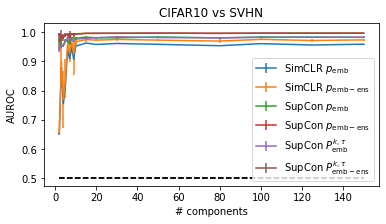}
     \end{subfigure}
     \begin{subfigure}{.48\textwidth}
         \includegraphics[width=\textwidth]{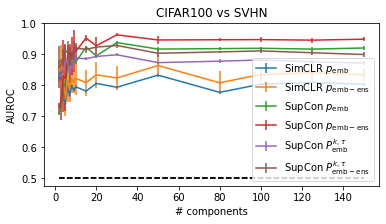}
     \end{subfigure}
     \begin{subfigure}{.48\textwidth}
         \includegraphics[width=\textwidth]{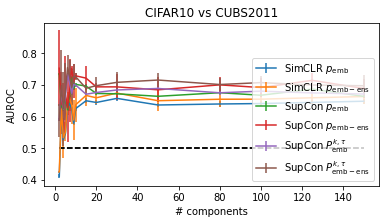}
     \end{subfigure}
     \begin{subfigure}{.48\textwidth}
         \includegraphics[width=\textwidth]{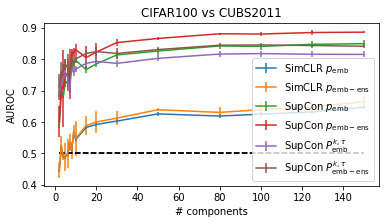}
     \end{subfigure}
     \begin{subfigure}{.48\textwidth}
         \includegraphics[width=\textwidth]{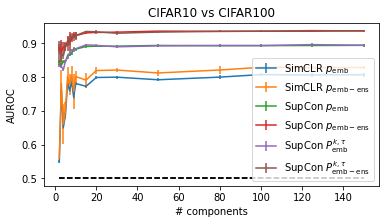}
     \end{subfigure}
     \begin{subfigure}{.48\textwidth}
         \includegraphics[width=\textwidth]{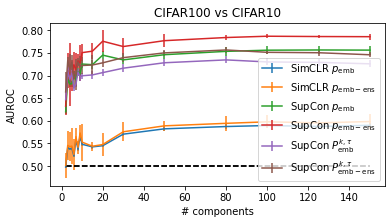}
     \end{subfigure}
        \caption{The effect of number of components in the GMM on different overall uncertainty measures. The titles specify the in-distribution (pre-training) datasets vs. the out-of-distribiution datasets. Similar to the other types of uncertainty, we generally observe improvements with larger $n_\text{comp}$. However the improvements are rather marginal for higher $n_\text{comp}$, and the results seem stable across all models and datasets.}
        \label{fig:overall_ncomp}
\end{figure}

Again, related to Section~\ref{sec:ablation}, we provide the effect of number of components on the overall uncertainty estimation, shown in Figure~\ref{fig:overall_ncomp}. We observe very similar trends compared to epistemic uncertainty, and we find the performance to be rather stable beyond $n_\text{comp} = 30$.  

\FloatBarrier

\end{document}


%

%

\onecolumn
\aistatstitle{Contrastive Uncertainty: \\
Supplementary Materials}












\vfill